\documentclass[conference]{IEEEtran}
\IEEEoverridecommandlockouts
\usepackage{cite}
\usepackage{amsmath,amssymb,amsfonts}
\usepackage{algorithmic}
\usepackage{graphicx}
\usepackage{textcomp}
\usepackage{xcolor}
\usepackage{hyperref}
\usepackage{float}
\def\BibTeX{{\rm B\kern-.05em{\sc i\kern-.025em b}\kern-.08em
    T\kern-.1667em\lower.7ex\hbox{E}\kern-.125emX}}
\begin{document}

\title{Indo LEGO-ABSA: A Multitask Generative Aspect Based Sentiment Analysis for Indonesian Language\\
% {\footnotesize \textsuperscript{*}Note: Sub-titles are not captured in Xplore and
% should not be used}
}

\author{\IEEEauthorblockN{Randy Zakya Suchrady}
\IEEEauthorblockA{\textit{School of Electrical Engineering and Informatics} \\
\textit{Institut Teknologi Bandung}\\
Bandung, Indonesia \\
23523027@mahasiswa.itb.ac.id}
\and
\IEEEauthorblockN{Ayu Purwarianti}
\IEEEauthorblockA{\textit{School of Electrical Engineering and Informatics} \\
\textit{Institut Teknologi Bandung}\\
Bandung, Indonesia \\
ayu@informatika.org}
}

\IEEEoverridecommandlockouts
\IEEEpubid{\makebox[\columnwidth]{979-8-3503-8129-0/23/\$31.00 \copyright2023 IEEE \hfill}
\hspace{\columnsep}\makebox[\columnwidth]{ }}

\maketitle

\begin{abstract}
Aspect-based sentiment analysis is a method in natural language processing aimed at identifying and understanding sentiments related to specific aspects of an entity. Aspects are words or phrases that represent an aspect or attribute of a particular entity. Earlier studies have applied generative pre-trained language model for aspect-based sentiment analysis. An example of this is the LEGO-ABSA framework, which effectively utilized these models, specifically in English-based aspect-based sentiment analysis. LEGO-ABSA uses a multitask learning and prompting approach to enhance model performance. However, the application of this approach has not been done in the context of Indonesian language. Therefore, this research aims to implement the multitask learning and prompting approach in aspect-based sentiment analysis for Indonesian language using generative pre-trained language model. In this study, the Indo LEGO-ABSA model is developed, which is an aspect-based sentiment analysis model utilizing generative pre-trained language model and trained with multitask learning and prompting. Indo LEGO-ABSA is trained with a hotel domain dataset in the Indonesian language. The obtained results include an f1-score of 79.55\% for the Aspect Sentiment Triplet Extraction, 86.09\% for Unified Aspect-based Sentiment Analysis, 79.85\% for Aspect Opinion Pair Extraction, 87.45\% for Aspect Term Extraction, and 88.09\% for Opinion Term Extraction. Indo LEGO-ABSA adopts the LEGO-ABSA framework that employs the T5 model, specifically mT5, by applying multitask learning to train all tasks within aspect-based sentiment analysis.\footnote{All works can be visited in \url{https://github.com/rdyzakya/IndoLEGO-ABSA}}
\end{abstract}


\section{Introduction}
Sentiment analysis, or opinion mining, investigates people's feelings and opinions about various entities like products, services, and events \cite{b1}. Analyzing sentiments is challenging due to ambiguous or domain-specific texts \cite{b2}. The analysis can be done at document, sentence, or phrase levels, with aspect-based sentiment analysis (ABSA) focusing on specific aspects \cite{b3}. ABSA operates at a lower linguistic level, aiming for detailed sentiment analysis \cite{b1}.
% Sentiment analysis, also known as opinion mining, is a field of study that analyzes opinions, sentiments, evaluations, judgments, attitudes, and emotions of people towards entities such as products, services, organizations, individuals, issues, events, topics, and their attributes \cite{b1}. Sentiment analysis is a challenging task due to the nature of texts categorized as having sentiments, which often lack clear meaning or require domain-specific knowledge to comprehend the sentiments within them \cite{b2}. Sentiment analysis is divided into document-level sentiment analysis, sentence-level sentiment analysis, and phrase-level sentiment analysis \cite{b3}. One form of phrase-level sentiment analysis is aspect-based sentiment analysis (ABSA). ABSA aims to conduct sentiment analysis at a lower linguistic level, namely the aspect level \cite{b1}.

In general, the issue of ABSA aims to identify four sentiment elements: aspect marker, aspect category, opinion marker, and sentiment polarity (or sentiment value) \cite{b4}\cite{b5}. The definitions of each sentiment element are as follows:
\newline

\noindent \textbf{Aspect Term} The target being described by opinion term, usually describes an attribute of an entity, it may appear explicitly or implicitly in the text \cite{b6}\cite{b7}.
\newline

\noindent \textbf{Aspect Category} A category of an aspect of an entity that falls into a specific category within a particular domain \cite{b8}.
\newline

\noindent \textbf{Opinion Term} An expression or statement that carry sentiment value towards a particular target \cite{b9}.
\newline

\noindent \textbf{Sentiment Polarity} A value representing the sentiment towards an aspect term or aspect category. Usually, sentiment values are positive, negative, and neutral \cite{b1}.

\begin{figure}[htbp]
\centerline{\includegraphics[scale=0.1]{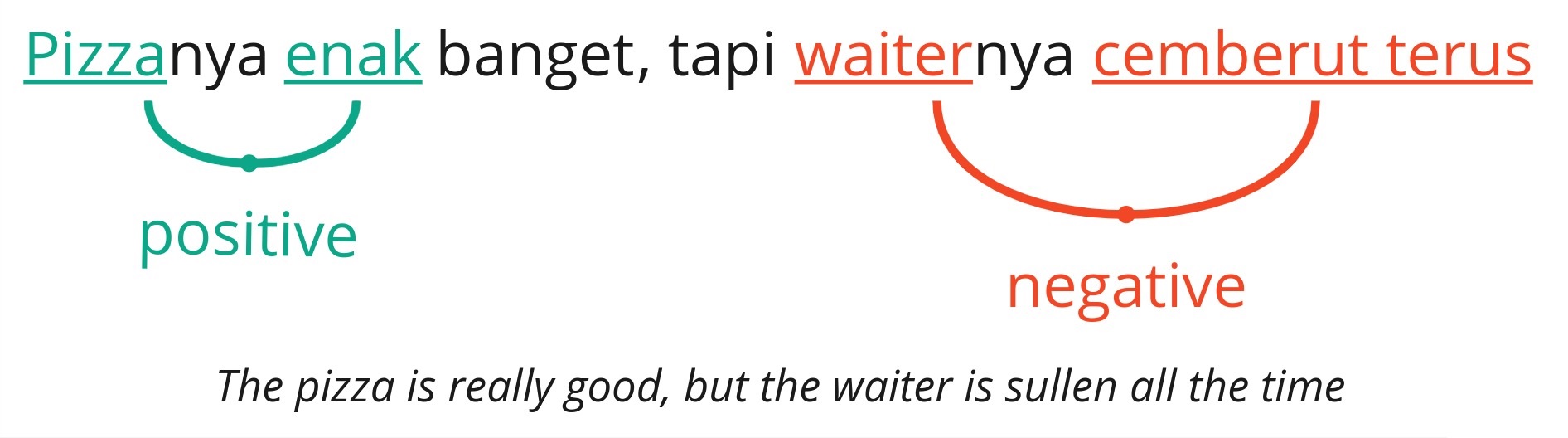}}
\caption{Illustration of aspect-based sentiment analysis in a sentence}
\label{fig1}
\end{figure}

For example, in Figure \ref{fig1}, the sentence has two aspect terms, namely "Pizza" and "waiter". The aspect term "Pizza" has a positive sentiment value indicated by the phrase "enak" (delicious) and the aspect term "waiter" has a negative sentiment value indicated by the phrase "cemberut terus" (sullen all the time). In the restaurant domain, the aspect term "Pizza" will be categorized as "food", and the aspect term
"waiter" will be categorized as "service".

ABSA comprises various tasks within it. There are Single Task, Basic Task, and Advance Task. Below are the details for each task:
\newline

\noindent \textbf{Single Task} Refers to tasks that aim to extract or classify only one sentiment element. Examples of single tasks include Aspect Term Extraction (ATE) \cite{b10}, Opinion Term Extraction (OTE) \cite{b11}, and Aspect Category Detection (ACD) \cite{b8}.
\newline

\noindent \textbf{Basic Task} Refers to tasks that aim to extract two sentiment elements simultaneously. Examples of these tasks are Aspect Opinion Pair Extraction (AOPE) \cite{b12}\cite{b13}, Unified Aspect Based Sentiment Analysis (UABSA) \cite{b14}\cite{b15}, and Aspect Category Sentiment Analysis (ACSA) \cite{b16}.
\newline

\noindent \textbf{Advance Task} Refers to tasks that aim to extract more than two sentiment elements. Examples of such tasks are Aspect Sentiment Triplet Extraction (ASTE) \cite{b9}, Target Aspect Sentiment Detection (TASD) \cite{b17}, and Aspect Category Opinion Sentiment Quadruple Extraction (ACOS) \cite{b18}.
\newline

In this paper, we propose \textbf{Indo LEGO-ABSA}, a generative ABSA model that tackles various tasks in ABSA especially for Indonesian language. We build the model based on the LEGO-ABSA framework.

\section{Related Works}
\subsection{Generative Aspect-based Sentiment Analysis (GAS)}
Previous research in the field of ABSA has predominantly followed a discriminative approach. Some studies have employed a sequence labeling methodology and have also leveraged multiple models for tasks requiring the extraction of multiple sentiment elements \cite{b9}\cite{b17}\cite{b19}\cite{b20}\cite{b21}\cite{b22}. However, this approach overlooks the semantics of labels, specifically with regard to aspect categories and sentiment polarity, and can lead to error propagation, especially when employing a pipeline with multiple models. This limitation in the existing approach served as motivation for Zhang et al. \cite{b23} to propose a unified generative framework known as Generative Aspect-based Sentiment Analysis (GAS). GAS adopts a sequence-to-sequence approach, taking a sentence as input and generating the corresponding sentiment elements as output. Two paradigms are employed for generating the answer: the extraction paradigm and the annotation paradigm. In the extraction paradigm, the answer is generated in the form of tuples separated by semicolons, while the annotation paradigm produces the answer in the form of the input sentence along with annotations. In the annotation paradigm, aspect terms are replaced with modified annotation marks enclosed in angle brackets. For instance, the sentence "The pizza is delicious" is annotated as "The [pizza $|$ delicious $|$ positive] is delicious". Notably, the results obtained from the GAS framework surpass those achieved by previous studies using discriminative approaches.
\subsection{GAS-Indonesia}
William \& Khodra \cite{b24} employed the GAS framework to address the Aspect Sentiment Triplet Extraction (ASTE) task within the context of the Indonesian language. To accommodate the Indonesian language, they utilized a modified version of the T5 model \cite{b25} specifically tailored for Indonesian language, known as IndoT5. The dataset used for this study originated from the hotel domain \cite{b26}, but with modifications involving the inclusion of implicit aspect terms. For these implicit aspect terms, the string "NULL" was implemented as a placeholder in answer generation. Throughout the experimentation, there were several augmentations made to the GAS framework, which encompassed elements like post-training, a novel normalization technique for optimizing method selection, and variations in the beam search process. %The findings indicate that the implementation of post-training negatively impacts the model's performance. It is advised to employ normalization techniques such as levenshtein distance and cosine similarity. Additionally, the optimal beam width was determined to be 1.
\subsection{LEGO-ABSA}
\begin{figure*}
\centerline{\includegraphics[scale=0.25]{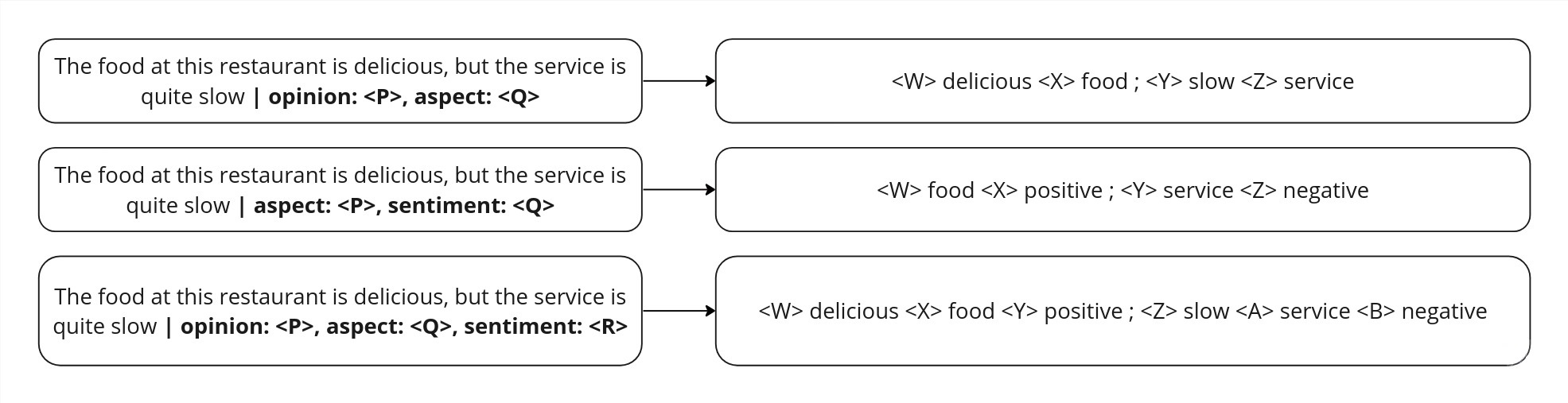}}
\caption{{Prompts and answers in LEGO-ABSA}}
\label{fig2}
\end{figure*}

In the previous study, each individual ABSA task was executed separately. Gao et al. \cite{b27} introduced a methodology that employed the T5 model, named LEGO-ABSA, to facilitate multitask learning across various ABSA tasks. This encompassed several ABSA tasks discussed in this investigation, specifically AOPE, UABSA (referred to as End to End Aspect Based Sentiment Analysis or E2E-ABSA), ACSA, ASTE, TASD, and ACOS (referred to as Aspect Sentiment Quad Prediction or ASQP). Beyond the application of multitask learning, additional concepts were introduced, including the assembly of task prompts and the notion of transfer tasks. Task prompt assembly involves a technique for constructing prompts tailored to the specific ABSA tasks. This involves combining prompts from simpler tasks and adapting them to suit more intricate tasks. As an example, prompts from UABSA and AOPE are employed as prompts for the ASTE task. Furthermore, there exists the transfer task, which entails the multitask training of the LEGO-ABSA model utilizing training data from simpler tasks, thus enabling the model to effectively handle more complex tasks. The construction of prompts and answers (can be seen in Figure \ref{fig2}) in LEGO-ABSA is inspired by the denoising technique employed by the T5 model during its unsupervised training phase. By providing prompts and answers familiar to T5, the hypothesis is that T5 doesn't need to put in as much effort during LEGO-ABSA training.

\section{Proposed Method}
While developing Indo LEGO-ABSA, we investigate five primary components: (1) the generative model, (2) the combination of training tasks, (3) the format of answers, (4) the format of prompts, and (5) supplementary datasets. We systematically progress through these components in sequence to identify the optimal candidate solution. This implies that the most effective solution from a preceding component will serve as the basis for exploring solutions in the subsequent component. It's important to note that our evaluation exclusively focuses on comparing candidate solutions based on their performance in the ASTE task. This choice is made because ASTE represents the most complex task among all the tasks under consideration.
\subsection{Generative Model}
%When selecting the potential solution for this particular element, we establish a set of criteria that must be met by each candidate model. These models need to be open-source generative models capable of accommodating the Indonesian language, while also avoiding any issues with memory overflow caused by the number of parameters. Based on these criteria, we opt for MBART (large) \cite{b28}, mT5 (base) \cite{b29}, and XGLM (564 million parameters) \cite{b30}. Concerning the combination of training tasks, we encompass all tasks in our training. For the answer format, we adopt the extraction paradigm present in GAS's answer format, which consists of sentiment element tuples separated by semicolons between them. Moreover, we employ a similar prompt structure as seen in LEGO-ABSA's prompt, where we include the answer format within the prompt itself.
We choose MBART (large) \cite{b28}, mT5 (base) \cite{b29}, and XGLM (564 million parameters) \cite{b30} as our model options. All training tasks are included. Our answer format follows GAS's pattern (extraction paradigm), with sentiment element tuples separated by semicolons. Additionally, we incorporate LEGO-ABSA's prompt structure, integrating the answer format directly within the prompt.

\subsection{Training Tasks Combination}
We restrict the combination of training tasks by grouping tasks together rather than considering each task individually. The training task combinations that arise from this approach encompass the basic task, advance task, single task + basic task, single task + advance task, basic task + advance task, and single task + basic task + advance task. We intentionally refrain from exploring the single task-only solution. This decision stems from the fact that individual single tasks fail to encompass all sentiment elements, which would result in the model's inability to recognize the complete array of sentiment elements. 

\subsection{Answer Format}

\begin{figure}[htbp]
\centerline{\includegraphics[scale=0.4]{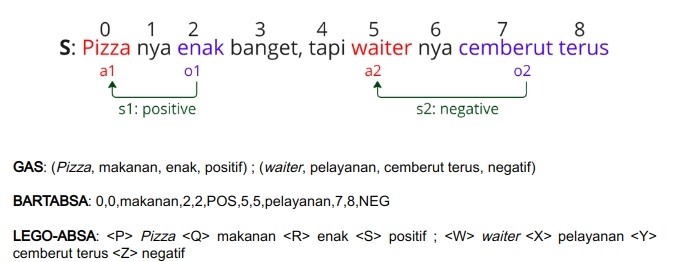}}
\caption{{Answer format candidate solution}}
\label{fig3}
\end{figure}

Regarding the answer format, we draw inspiration from prior research. We consider several options for the answer format, including the GAS \cite{b23} answer format (extraction paradigm), the BARTABSA \cite{b31} answer format, and the LEGO-ABSA \cite{b27} answer format.

\subsection{Prompt Format}

\begin{figure}[htbp]
\centerline{\includegraphics[scale=0.21]{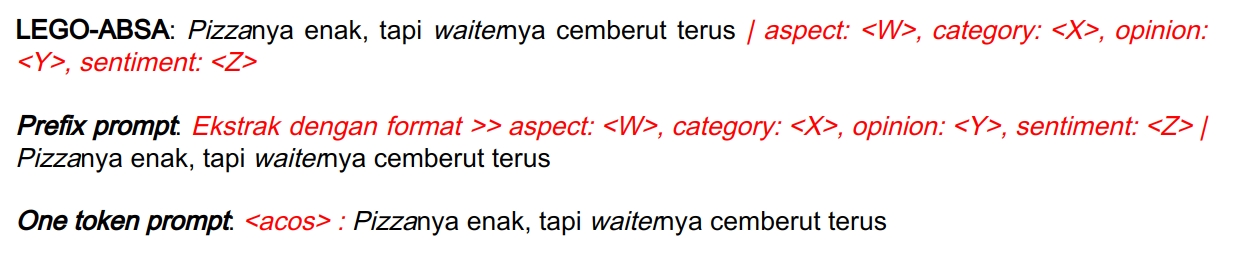}}
\caption{{Prompt format candidate solution}}
\label{fig4}
\end{figure}

Previous research has demonstrated that only LEGO-ABSA has incorporated the use of prompts. This is attributed to the multitask setup, which enables the model to distinguish between different tasks. In this study, we also introduce an alternative prompt format, consisting of both the prefix prompt and the one token prompt. The prefix prompt resembles the LEGO-ABSA prompt, with the addition of specific instructions and its placement at the beginning of the input text. On the other hand, the one token prompt employs a single token to represent each task, as opposed to a sequence of tokens.

\subsection{Supplementary Datasets}
We aim to extend the scope of the multitask setting beyond ABSA tasks alone. We incorporate additional supplementary datasets to investigate whether they can contribute to improving the performance of solving ABSA tasks. Specifically, we select three tasks: part-of-speech tagging (POS tagging), document-level sentiment analysis, and emotion classification.

\section{Experiment}

\subsection{Dataset}
In this paper we adopt the same dataset in GAS Indonesia. The dataset used is a collection of Indonesian language data in the hotel domain, consisting of 5000 texts (3000 train, 1000 validation, 1000 test) taken from Wirawan's research \cite{b26}. This dataset comprises only one main task, which is ASTE. The data was modified by adding implicit aspect terms. In triplets with implicit aspect terms, the aspect terms are not explicitly mentioned in the sentence, but only the opinion terms are present. The addition of triplets with implicit aspect terms is based on observations in daily life. Many studies use data that only have explicit aspect terms without the presence of implicit aspect terms. Reviews of products, as one of the subjects in ABSA research, often have implicit aspect terms \cite{b18}. With this basis, William \& Khodra \cite{b24} added triplets with implicit aspect terms. The string "NULL" is chosen to represent the implicit aspect term. In order to establish the multitask setting, we transform the dataset initially focused solely on the ASTE task into multiple distinct tasks, namely ATE, OTE, AOPE, and UABSA.

\subsection{Implementation Details}

\noindent\textbf{Evaluation Metrics} The metrics employed in this study adhere to prior research practices. We utilize the f1-score metrics, wherein a tuple is considered true only if all the sentiment elements within it are accurately predicted.
\newline

\noindent\textbf{Experiment Details} In the searching of the best candidate solution, we traverse the components sequentially. For each candidate solution, we maintain uniformity in the hyperparameters for all stages, employing 10 epochs, a learning rate of 3e-4, a batch size of 8, and 2 gradient accumulation steps.

\subsection{Main Result}
We obtained candidate solution outcomes using mT5 as the foundational model, training all tasks (single task + basic task + advance task), LEGO-ABSA answer format, prefix prompt format, and integrated POS tagging into the training dataset. This outcome was then contrasted with the vanilla LEGO-ABSA, which was trained exclusively on Indonesian language (mT5, all tasks trained, LEGO-ABSA answer format, LEGO-ABSA prompt format, and no additional dataset). Interestingly, during the comparative analysis, the vanilla LEGO-ABSA demonstrated a higher f1-score. The detailed outcomes are presented in Table \ref{tab1}.

\begin{table}[H] %[htbp]
\caption{Best candidate solution comparison}
    \begin{center}
        \begin{tabular}{|c|c|c|c|c|c|}
        \hline
              & \textbf{ASTE} & \textbf{UABSA} & \textbf{AOPE} & \textbf{ATE} & \textbf{OTE} \\
        \hline
        Exploration Result & \textbf{80.31} & 85.63 & 79.55 & 86.75 & 87.67 \\
        \hline
        \textbf{Vanilla LEGO-ABSA} & 79.55 & \textbf{86.09} & \textbf{79.85} & \textbf{87.45} & \textbf{88.09} \\
        \hline
        \end{tabular}
    \end{center}
    \label{tab1}
\end{table}

We label the outcome of the comparison as \textbf{Indo LEGO-ABSA}. Subsequently, we compare \textbf{Indo LEGO-ABSA} with GAS Indonesia (using mT5 as the base model due to fairness). To facilitate this comparison, we develop a GAS Indonesia model for each task, employing the extraction paradigm format. The comparison result can be seen in Table \ref{tab2}.

\begin{table}[H] %[htbp]
\caption{Indo LEGO-ABSA vs GAS Indonesia}
    \begin{center}
        \begin{tabular}{|c|c|c|c|c|c|}
        \hline
              & \textbf{ASTE} & \textbf{UABSA} & \textbf{AOPE} & \textbf{ATE} & \textbf{OTE} \\
        \hline
        GAS Indonesia & 78.63 & 82.34 & 78.45 & 84.02 & 87.74 \\
        \hline
        \textbf{Indo LEGO-ABSA} & \textbf{79.55} & \textbf{86.09} & \textbf{79.85} & \textbf{87.45} & \textbf{88.09} \\
        \hline
        \end{tabular}
    \end{center}
    \label{tab2}
\end{table}

We didn't notice any big changes in how the model makes inferences compared to earlier results. The only difference is that the model performs better when using multitask training. This is because the tasks benefit from each other's inductive biases, which leads to improved performance.

\subsection{Error Analysis}
We extracted 30 instances of inference results from the ASTE task that exhibited errors. These errors encompassed approximately 10 prevalent patterns. We classify these patterns as ANNOTATION (6 occurrences), UNDERPERFORM (16 occurrences), IMPLICIT (6 occurrences), INCOMPLETE (4 occurrences), POS\textunderscore CONFUSE (3 occurrences), SERIES (3 occurrences), TYPO (11 occurrences), SENTENCE\textunderscore STRUCTURE (2 occurrences), COREFERENCE (1 occurrence), and TRAIN\textunderscore DATA (1 occurrence).

\begin{table}[H] % [htbp]
% \caption{Annotation Error}
    \centering
    \begin{tabular}{|l|l|}
    \hline
    \multicolumn{1}{|c|}{\textbf{Text}} & \multicolumn{1}{|c|}{\textbf{Target}} \\
    \hline
    tidak terlalu jauh dari pusat kota & (dari pusat kota, tidak \\
     dan ada coffee yang super enak .& terlalu jauh, POS)\\
    \hline
    \end{tabular}
    \label{tab3}
\end{table}

\noindent \textbf{ANNOTATION} Refers to the errors that occurred originated from annotation mistakes in the test data. In the above example, the proposed aspect term in the annotation, "dari pusat kota" (from the city center) does not meet the definition of an aspect term, as "tidak terlalu jauh" (not too far) does not describe "dari pusat kota" but rather describes an aspect term that is not present in the text.

\begin{table}[H] %[htbp]
% \caption{Underperform Error}
    \centering
    \begin{tabular}{|l|l|l|}
    \hline
    \multicolumn{1}{|c|}{\textbf{Text}} & \multicolumn{1}{|c|}{\textbf{False Positive}} &  \multicolumn{1}{|c|}{\textbf{False Negative}}\\
    \hline
    bagus dan bersih . & (NULL, bagus dan, POS)& (NULL, bagus, POS);\\
     & & (NULL, bersih, POS)\\
    \hline
    \end{tabular}
    \label{tab4}
\end{table}

\noindent \textbf{UNDERPERFORM} The errors occurred due to the ABSA model's inability to perform extraction on the related tuples. In the above example, the word "dan" (and) is inserted into the extracted aspect term, and this cannot be explained further. Another tuple with the opinion term "bersih" (clean) also failed to be extracted by the model.

\begin{table}[H] %[htbp]
% \caption{Implicit Error}
    \centering
    \begin{tabular}{|l|l|}
    \hline
    \multicolumn{1}{|c|}{\textbf{Text}} & \multicolumn{1}{|c|}{\textbf{False Positive}}\\
    \hline
    aku suka disini . dapat kue basah & (kue basah, dapat, POS)\\
    dan ada coffee yang super enak . & \\
    \hline
    \end{tabular}
    \label{tab5}
\end{table}

\noindent \textbf{IMPLICIT} The errors occurred because the ABSA model struggled to capture implicitly conveyed sentiments. In the above example, the phrase "dapat kue basah" (get moist cake) is considered as something positive. There is a possibility that the model perceives that there is an implicitly conveyed sentiment from "dapat kue basah".
\newline

\begin{table}[H] %[htbp]
% \caption{Incomplete Error}
    \centering
    \begin{tabular}{|l|l|l|}
    \hline
    \multicolumn{1}{|c|}{\textbf{Text}} & \multicolumn{1}{|c|}{\textbf{Target}} & \multicolumn{1}{|c|}{\textbf{False Positive}} \\
    \hline
    ...tanpa ada lift . & (lift, tanpa ada, NEG)& (lift, tanpa, NEG)\\
    \hline
    \end{tabular}
    \label{tab6}
\end{table}

\noindent \textbf{INCOMPLETE} The extracted tuples are correct; however, there are parts of the words from the aspect term or opinion term that are incomplete. In the above example, the model failed to extract the word "ada" (there is), which is a continuation of the word "tanpa" (without).
\newline

\begin{table}[H] %[htbp]
% \caption{Pos\textunderscore Confuse Error}
    \centering
    \begin{tabular}{|l|l|l|}
    \hline
    \multicolumn{1}{|c|}{\textbf{Text}} & \multicolumn{1}{|c|}{\textbf{Target}} & \multicolumn{1}{|c|}{\textbf{False Positive}} \\
    \hline
    cocok untuk & (NULL, cocok, POS);& (guest house, cocok\\
    keluarga , seperti & (NULL, bersih, POS)& , POS); (guest house,\\
    layaknya guest & & bersih POS)\\
    house , bersih . & & \\
    \hline
    \end{tabular}
    \label{tab7}
\end{table}

\noindent \textbf{POS\textunderscore CONFUSE} The ABSA model's inability arises when within a tuple, the aspect term and opinion term have ambiguous part of speech. In the above example, the word "cocok" (suitable) should refer to the aspect term described as "seperti layaknya guest house" (like a guest house) and not to the "guest house" itself. This could happen because "guest house" is the nearest noun that matches the context of "cocok". Ideally, the entire description "seperti layaknya guest house" is a phrase that indicates the similarity of the implicit aspect term being discussed with the "guest house".
\newline

\begin{table}[H] %[htbp]
% \caption{Series Error}
    \centering
    \begin{tabular}{|l|l|}
    \hline
    \multicolumn{1}{|c|}{\textbf{Text}} & \multicolumn{1}{|c|}{\textbf{False Positive}} \\
    \hline
    ...full fasilitas . mulai air & (air hangat, lengkap, POS);\\
    hangat , ac , dan tv dengan & (ac, lengkap, POS);\\
    saluran lengkap . & (tv, lengkap, POS)\\
    \hline
    \end{tabular}
    \label{tab8}
\end{table}

\noindent \textbf{SERIES} The errors made by the ABSA model occur when a sequence of aspect terms are indicated by a single opinion term or when an aspect term is indicated by a sequence of opinion terms, and the model fails to extract them correctly. In the above example, "air hangat" (warm water) and "ac" (air conditioning) are included in the sequence of objects that end with the word "lengkap" (complete), which actually refers to "saluran tv" (TV channels).
\newline

\begin{table}[H] %[htbp]
% \caption{Typo Error}
    \centering
    \begin{tabular}{|l|l|}
    \hline
    \multicolumn{1}{|c|}{\textbf{Text}} & \multicolumn{1}{|c|}{\textbf{False Positive}} \\
    \hline
    yang paling saya suka ada smoking . & (smoking areaanya, ada, POS)\\
    areanya . & \\
    \hline
    \end{tabular}
    \label{tab9}
\end{table}

\noindent \textbf{TYPO} The errors in aspect term or opinion term writing are made by the ABSA model or errors in writing within the sentence. In the above example, the model has extracted correctly; however, there is a difference in the way the word "areanya" (the area) is written, obtained by the model as "areaanya".
\newline

\begin{table}[H] %[htbp]
% \caption{Sentence\textunderscore Structure Error}
    \centering
    \begin{tabular}{|l|l|l|}
    \hline
    \multicolumn{1}{|c|}{\textbf{Text}} & \multicolumn{1}{|c|}{\textbf{Target}} & \multicolumn{1}{|c|}{\textbf{False Positive}} \\
    \hline
    wifi nya tolong& (wifi nya, tolong& (wifi nya, tolong\\
 diperkuat sinyal dan& diperkuat, NEG)&diperkuat sinyal,\\
     ditambah bandwithnya .& & NEG)\\\hline
    \end{tabular}
    \label{tab10}
\end{table}

\noindent \textbf{SENTENCE\textunderscore STRUCTURE} Errors by the ABSA model are caused by ambiguous or messy sentence structures. In the above example, the model made an error in extracting the opinion term "tolong diperkuat" (please strengthen) as "tolong diperkuat sinyal" (please strengthen signal). This can happen due to the poor sentence structure where the word "sinyal" (signal) can be positioned before "wifi" or omitted. This indicates that the model encounters difficulties when the meaning of the input sentence becomes ambiguous due to non-standard sentence structures.
\newline

\begin{table}[H] %[htbp]
% \caption{Coreference Error}
    \centering
    \begin{tabular}{|l|l|l|}
    \hline
    \multicolumn{1}{|c|}{\textbf{Text}} & \multicolumn{1}{|c|}{\textbf{Target}} & \multicolumn{1}{|c|}{\textbf{Inference Result}} \\
    \hline
    ...yang bisa dipakai & (yang bisa dipakai, & -\\
    cuma 1 lubang... & cuma 1, negative) & \\
    \hline
    \end{tabular}
    \label{tab11}
\end{table}

\noindent \textbf{COREFERENCE} The errors made by the ABSA model occur when the aspect term is a pronoun that refers to the actual aspect term. In the above example, the aspect term "yang bisa dipakai" (that can be used) is a phrase that refers to the usable "lubang" (refer to electrical socket). This indicates that phrases that refer to another aspect term are difficult to be extracted by the model.
\newline

\begin{table}[H] %[htbp]
% \caption{Train\textunderscore Data Error}
    \centering
    \begin{tabular}{|l|l|l|}
    \hline
    \multicolumn{1}{|c|}{\textbf{Text}} & \multicolumn{1}{|c|}{\textbf{Target}} & \multicolumn{1}{|c|}{\textbf{Inference Result}} \\
    \hline
    teko air , meja , & - & (teko air, masih, POS);...\\
    peralatan lainnya . & & \\
    \hline
    \end{tabular}
    \label{tab12}
\end{table}

\noindent \textbf{TRAIN\textunderscore DATA} The errors of the ABSA model are caused by the lack of sentence variations in the training dataset with similar characteristics. In the above example, there are no tuples that need to be extracted by the model, but the model erroneously performs extraction on the tuples. This can happen because there is a small number of sentence variations in the training data without target tuples. This is further supported by the finding that in the training data, there are only 164 out of 3000 texts without target tuples.
\newline

% To improve the model’s performance, we try to do a deeper analysis towards errors with the category TYPO. TYPO errors usually can be repaired by a normalization algorithm that reconstructs the falsely generated words. In Table \ref{tab13}, we can see some samples from the TYPO error.

% \begin{table}[H] %[htbp]
% \caption{Typo Error Variation}
%     \centering
%     \begin{tabular}{|l|l|l|}
%     \hline
%     \multicolumn{1}{|c|}{\textbf{Text}} & \multicolumn{1}{|c|}{\textbf{Generated Word}} & \multicolumn{1}{|c|}{\textbf{Reference Word}} \\
%     \hline
%     ...suka ada smoking& "areaanya" & "areanya" \\
%     areanya .& & \\
%     \hline
%     hotelnya biasa & "biasa biasa& "biasa biasa aja"\\ 
%     biasa saja...& biasa biasa aja"& \\
%     \hline
%     furniture baiknya & "furniture & "kamar mandinya"\\ 
%     ...kamar mandinya...& mandinya" & \\
%     \hline
%     ...sesuai dengan& "sesuai harganya" & "sesuai dengan\\
%  harganya .& &harganya"\\\hline
%     \end{tabular}
%     \label{tab13}
% \end{table}

% \noindent The discovered generation errors vary, ranging from differences in character levels, repetitive generated words, the appearance of new phrases, to the omission of middle words. The diverse variations of these errors make it difficult to find the right algorithm for normalization. Based on this, normalization was not conducted in this research.
% \newline

From the error analysis, we do a further investigation through the training data to see if there are any patterns that leads to the errors. We conducted a sampling of 100 text instances from our training dataset and found some interesting insights. We observed that certain sentence structures posed challenges for interpretation, even for human annotators. Specifically, declarative sentences, characterized by their straightforward statements, presented difficulties in discerning the presence of sentiments within the text. Additionally, some text segments exhibited an implicit nature, wherein statements or suggestions subtly conveyed sentiments. For instance, consider the sentence "tempat parkir perlu lebih rapi" (the parking lot needs to be neater), which implies another sentence with an explicit characteristic, "tempat parkir tidak rapi" (the parking lot is not neat). This subtle distinction introduced inconsistency in the annotation process. While human annotators correctly annotated certain challenging sentences (e.g., declarative sentences or suggestions), they occasionally struggled with texts sharing similar characteristics. Moreover, the informal style of the texts posed additional complexities for our model's inference. Informal texts frequently exhibited ambiguous sentence structures and contained numerous typographical errors. These typos resulted in divergent tokenization and, consequently, divergent inference outcomes. Among the 100 sampled instances from the training dataset, our analysis revealed that only 41\% exhibited no defects, encompassing declarative sentences, implicit sentences, typos, or informal language usage.

\section{Conclusion}
Based on the conducted experiments and analyses, we have developed a model for ABSA named \textbf{Indo LEGO-ABSA}. The comprehensive conclusions drawn from this research are as follows: (1) The optimal generative pre-trained language model for aspect-based sentiment analysis in Indonesian, capable of multitask learning and prompting, is the mT5 model, benefiting from its familiarity with these techniques. (2) The most effective task combination for training a generative model for aspect-based sentiment analysis in Indonesian involves incorporating all task levels, including single, basic, and advanced tasks. (3) The ideal answer format is the LEGO-ABSA format, while the preferred prompt format is LEGO-ABSA prompt, both of which align with the mT5 model's familiarity with these formats.

\section*{Future Works}
There is room for further advancement in studies concerning aspect-based sentiment analysis in the Indonesian language. Several recommendations for future research include: (1) Exploring aspect-based sentiment analysis tasks that encompass aspect categories, considering that this study only involved three sentiment elements. (2) Investigating effective approaches for addressing errors in answer generation through generative models for aspect-based sentiment analysis in Indonesian language. Additionally, delving deeper into normalization algorithms could enhance the overall performance of the aspect-based sentiment analysis model.

\end{document}